\documentclass{article}

\PassOptionsToPackage{numbers, compress}{natbib}

\usepackage[final]{mlsb}

\usepackage[utf8]{inputenc} 
\usepackage[T1]{fontenc}    
\usepackage{hyperref}       
\usepackage{url}            
\usepackage{booktabs}       
\usepackage{amsfonts}       
\usepackage{nicefrac}       
\usepackage{microtype}      
\usepackage{xcolor}         
\usepackage{hyperref}
\usepackage{graphicx}
\graphicspath{ {./figures/} }
\graphicspath{{figuers/}}

\usepackage{epstopdf}
\DeclareGraphicsExtensions{.png,.pdf}

\title{ContactNet: Geometric-Based Deep Learning Model for Predicting Protein-Protein Interactions  }

\author{%
  Matan Halfon \\
  School of Computer Science and Engineering\\
  The Hebrew University of Jerusalem\\
  Jerusalem, Israel \\
  \texttt{matan.halfon@mail.huji.ac.il} \\
  \And
  Tomer Cohen \\
  School of Computer Science and Engineering\\
  The Hebrew University of Jerusalem\\
  Jerusalem, Israel \\
  \texttt{tomer.cohen13@mail.huji.ac.il} \\
 \And
    Raanan Fattal \\
  School of Computer Science and Engineering\\
  The Hebrew University of Jerusalem\\
  Jerusalem, Israel \\
  \texttt{raanan.fattal@mail.huji.ac.il} \\
  \And
  Dina Schneidman-Duhovny \\
  School of Computer Science and Engineering \\
  The Hebrew University of Jerusalem\\
  Jerusalem, Israel \\
  \texttt{dina.schneidman@mail.huji.ac.il } \\
}

\begin{document}

\maketitle
\begin{abstract}
Deep learning approaches achieved significant progress in predicting protein structures. These methods are often applied to protein-protein interactions (PPIs) yet require Multiple Sequence Alignment (MSA) which is unavailable for various interactions, such as antibody-antigen. Computational docking methods are capable of sampling accurate complex models, but also produce thousands of invalid configurations. The design of scoring functions for identifying accurate models is a long-standing challenge. We develop a novel attention-based Graph Neural Network (GNN), ContactNet, for classifying PPI models obtained from docking algorithms into accurate and incorrect ones. When trained on docked antigen and modeled antibody structures, ContactNet doubles the accuracy of current state-of-the-art scoring functions, achieving accurate models among its Top-10 at 43\% of the test cases. When applied to unbound antibodies, its Top-10 accuracy increases to 65\%. This performance is achieved without MSA and the approach is applicable to other types of interactions, such as host-pathogens or general PPIs.


\end{abstract}

\section{Introduction}
Experimental methods for structure determination of macromolecular complexes, such as X-ray crystallography and cryo- Electron Microscopy, can not be applied in a high-throughput manner \cite{cheng2015single}. Recent progress in deep learning models for protein folding also enabled modeling of protein-protein complexes with high accuracy \cite{baek2021accurate,evans2021protein,jumper2021highly}. However, these methods rely on a co-evolutionary signal from MSA which is not available for many complexes, such as antibody-antigen, leading to inaccurate predictions \cite{yin2022benchmarking}. Computational docking methods for modeling PPIs, while applicable on a larger scale, still suffer from low accuracy {\cite{lensink2018challenge}}. Docking usually involves two parts: sampling and scoring. Sampling methods, such as Fast Fourier Transform (FFT) or geometry-based approaches \cite{ritchie2000protein,gray2003protein,kozakov2006piper}, usually sample accurate models for the “rigid” docking cases, where there are no significant conformational changes between the bound and unbound structures. However, the scoring presents a major challenge in identifying a few accurate models among the thousands that are generated by the sampling methods.

Much effort has been devoted in developing scoring functions for assessing docked complexes \cite{comeau2004cluspro,andrusier2007firedock,renaud2021deeprank}. Despite this effort, the accuracy remains fairly low: state-of-the-art scoring functions rank correct models among its Top-10 models in 10\% to 40\% of the cases for unbound docking \cite{guest2021expanded}, depending on the dataset and the scoring function. The accuracy drops significantly, below 15\%, when the comparative models are used for docking \cite{singh2020application}. Recently, there has been progress in designing scoring functions using deep learning models, including 3D convolutional \cite{wang2020protein} and hierarchical rotation-equivariant networks \cite{eismann2021hierarchical}. However, they were not able to surpass traditional scoring functions, such as ZRANK \cite{zrank} and SOAP-PP \cite{dong2013optimized}. Graph representation and GNN models are a well-suited for protein structures and protein-protein interactions resulting in some success in scoring tasks \cite{dai2021protein,wang2021protein}.  
Here, we describe a new GNN-based model, ContactNet, which is specifically designed to model biological protein-protein recognition interfaces consisting of short interacting linear fragments with the appropriate invariances. The model detects and analyzes contacting patches in protein-protein interfaces to resolve the docked model validity. Specifically, the network consists of the following processing stages. First, it encodes an embedding of both the structural (geometric) and chemical properties of each residue in each input protein. The geometry is introduced by a variant of graph attention mechanism that limits the interaction to spatially neighboring amino acids. Second, the inter-protein distance matrix is used for detecting potentially interacting linear segments, which are also analyzed and encoded into low-dimensional contact descriptor vectors. Finally, a sequence of encoding-transformer layers integrates the information gathered from all the interacting linear segments to score input docked models. ContactNet was trained and tested on antibody-antigen data and achieved highly accurate classifications of antibody-antigen models generated by docking, where it nearly doubled the success rate of state-of-the-art methods. Specifically, it detected an accurate model among the Top-10 best scoring for $\sim$43\% of the test set, compared to $\sim$22\% by current methods.

\section{Methods}

\subsection{ContactNet architecture} ContactNet  starts by extracting low-level features at the level of neighboring amino acids in each protein and uses them to analyze larger scale contact regions and resolve their global interaction across the interface between the two proteins. This processing pipeline is performed by a sequence of three modules (Fig. \ref{cnet}):

\begin{figure}
\centering
\includegraphics[width=\textwidth]{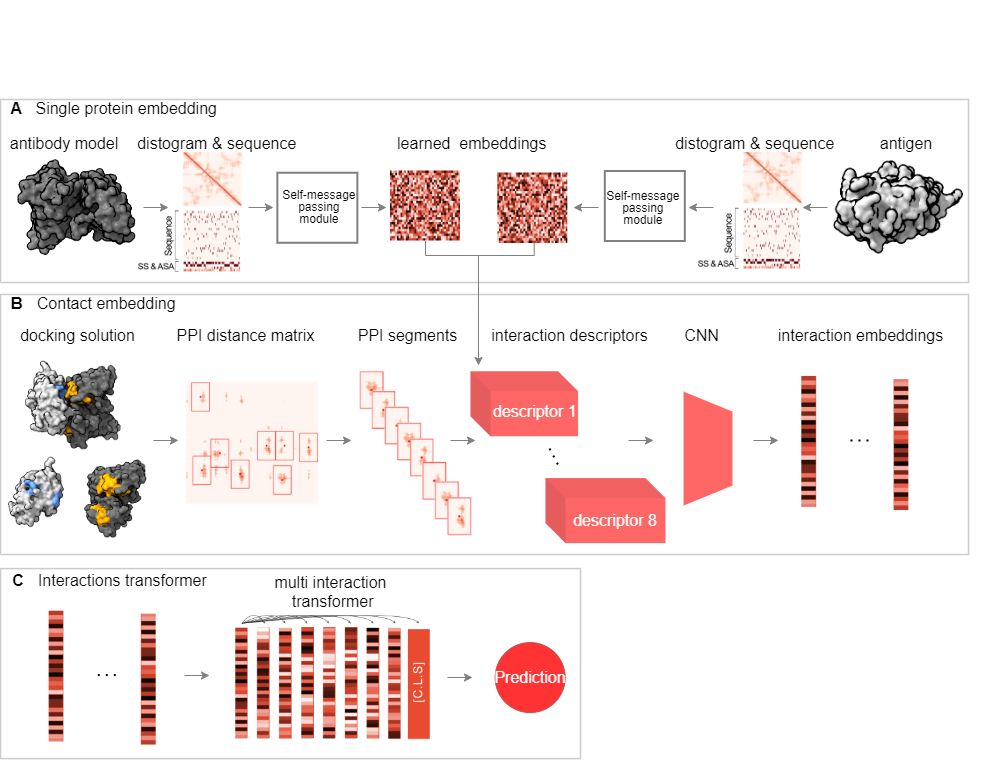}
\caption{ \textbf{ContactNet architecture.} \textbf{A} Single protein embedding module. This module uses sequence, secondary structure, solvent accessibility, and the $C\alpha-C\alpha$ protein distance matrix. The network learns a new representation of physico-chemical features using an encoder transformer. \textbf{B} Contacts embedding module. This module extracts the most contacting and possibly interacting linear segments from the single protein embedding stage and encodes them into interaction descriptors. \textbf{C} Interaction transformer module. The transformer module is trained to classify the entire interaction interfaces based on these embedded contact descriptors.}
\label{cnet}
\end{figure}
	
\textbf{Single-protein embedding module } This module uses the chemical properties of each amino acid in each protein along with its neighboring amino acids and contextualizes this information in an embedded representation. We implement this module using a distance aware graph attention layer \cite{velivckovic2017graph} where $C\alpha$ atoms are the nodes and the edges correspond to amino acids in  close proximity ($C\alpha-C\alpha$ distance < 16\AA). Specifically, we calculate the attention weight between two residues as $w_{ij}=softmax(<k_{i},q_j>/d_{ij})$ where $k_i=MLP(r_i)$ and  $q_j=MLP(r_j)$ for residue $i$, and obtain a context-aware representation by $e_i= \sum_{0}^{N}w_{i,j}v_j$.  allows us to introduce an inductive bias that decouples non-neighboring residues. The embedded representation is computed for each protein in the complex separately and is denoted by $e^{receptor}_{i}$ and $e^{ligand}_{i}$.

\textbf{Contacts embedding module}  The association between the proteins (binding or not) is mainly determined by the type of interaction between localized linear segments. Typically, there are around 6 peptide segments in an average interface of 1,000\AA$^2$ with most of the segments up to 13 amino acids in length \cite{pal2007peptide,london2010can}. We carefully design our network architecture to induce its inference according to these structural considerations.

The inter-protein distogram $D_{ij}$ (distance matrix of the receptor-ligand complex bipartite-graph) is a fairly sparse, and its non-trivial elements correspond to contacts between the two proteins. Consequently, we focus our classification on the neighboring segments that are identified using a non-maximum-suppression algorithm  over the inter-protein distogram matrix $D_{ij}$. We then extract short linear segments around these amino acids (10 amino acids in each chain direction) from their embedded single-protein representation, i.e., $[e^{receptor}_{i-L},...,e^{receptor}_{i+L}]$ and $[e^{ligand}_{j-L},...,e^{ligand}_{j+L}]$, where the length of these segments is 2L+1. To allow the network to account for the interaction between every pair of amino acids in the two segments, we generate an interaction descriptor (Fig. \ref{cnet}B) by tiling each segment along with the other and concatenating the two, as given by $E_{i,j}(s,t)= [e^{receptor}_{i+s} ; e^{ligand}_{j+t}]$ {\cite{wang2017accurate}}. Thus, the spatial coordinates $s,t$ of the resulting 3D tensor, $E_{i,j}(s,t)$, correspond to different pairs of amino acids, and its third coordinate parametrizes their embedded representation. By feeding each $E_{i,j}(s,t)$ through a 2D convolutional network, containing multiple pooling operators, we allow the model to analyze multiple sub-segments at different scales and lengths and produce a low-dimensional encoding vector $c_{ij}$ for each i,j contact. This process is applied over the top 8 non-overlapping interacting segment pairs detected from the inter-protein distogram $_{Dij}$.

\textbf{Interaction transformer module} The resulting contact descriptor vectors $c_{ij}$ are then passed, as separate tokens, to a classification transformer network (along with a special classification token). This global operator allows the model to integrate data from all the contacting segments to derive a final prediction of the validity of the complex. This module is capable of learning general non-linear relations between the contacts while being invariant to their order. The resulting classification token is then passed through an MLP to produce the final prediction.

\section{Results}
\begin{figure}
\centering
\includegraphics[width=\textwidth]{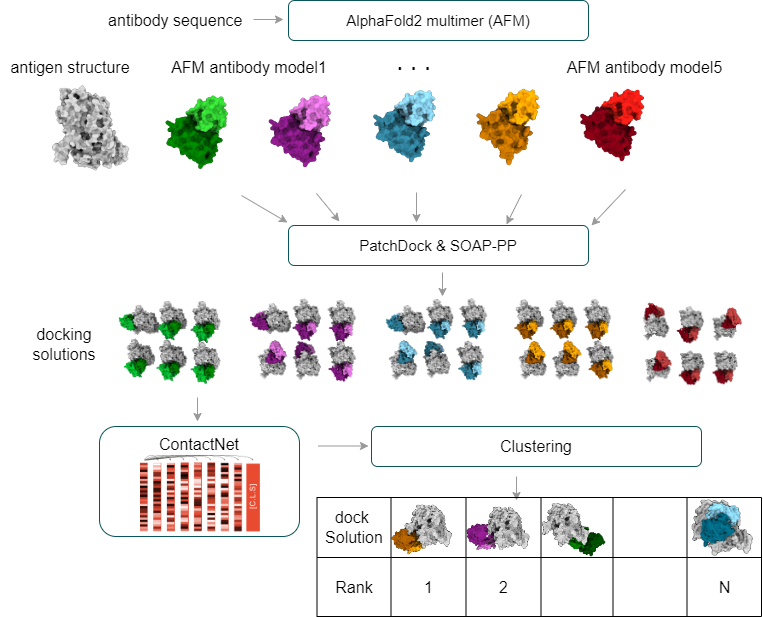}
\caption{Prediction protocol for antigen-antibody complexes. The antibody sequence is modeled via AFM to obtain five modeled structures that are docked using PatchDock. The top docking models ranked by SOAP-PP score are reevaluated by ContactNet. Finally, the docking models from the five AFM antibody models are clustered to reduce redundancy.
} 
\label{pipeline}
\end{figure}

\textbf{Prediction protocol for antigen-antibody complexes.} We trained and tested ContactNet on 875 antibody-antigen structures from the AbDb dataset (see \ref{Dataset}). The antibodies were modeled ab initio using AlphaFold2-multimer-v1 (AFM) \cite{evans2021protein} to simulate real-life scenario where only antibody sequences are available. Overall, five models were obtained for most of the antibodies. Docking was run on the antigen and five antibody AFM models using the antibody-antigen PatchDock protocol (see \ref{sampling}, Fig. \ref{pipeline}). Docking models were re-ranked using SOAP-PP statistical potential \cite{dong2013optimized} and the top 3000 were assessed by ContactNet. Finally, we clustered the docking models among the five antibody AFM models using interface clustering \cite{comeau2004cluspro} with an RMSD threshold of 5\AA. 

\textbf{ContactNet performance.} Overall, due to the sampling protocol (antibody modeling and docking), acceptable or higher accuracy docking models according to CAPRI criteria (see \ref{asessment}) were obtained for 88\% of the complexes (out of 85 in the test set), which is the upper bound for our success rate (Fig. \ref{results}A). Most of the correct models produced by docking were of acceptable and medium accuracy, with high accuracy models sampled rarely. We compared ContactNet to SOAP-PP statistical potential which defines the state-of-the-art on several benchmarks  and to the five antibody-antigen complex models obtained from the AFM, as it was shown to outperform state-of-the-art docking methods on several benchmarks \cite{evans2021protein,yin2022benchmarking}.

\begin{figure}
\centering
\includegraphics[scale=0.5]{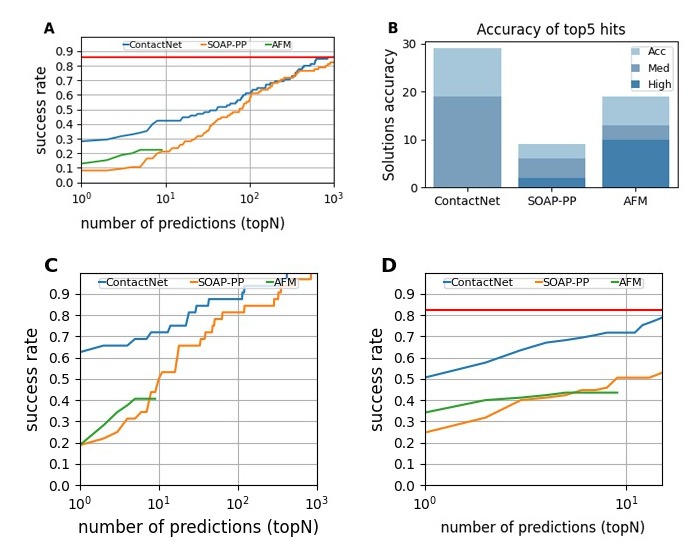}
\caption{\textbf{Performance of ContactNet, AFM, and SOAP-PP on the modeled antibodies test set.} \textbf{A.} Success rate for TopN predictions for ContactNet (blue), AFM (green), and SOAP-PP (orange). The red line indicates the upper bound for ContactNet and SOAP-PP due to missing acceptable or higher accuracy models.\textbf{B.} Top5 models of each method divided into high, medium, and acceptable accuracy. 
\textbf{C.} Success rate for topN predictions on the unbound test set for ContactNet, AFM, and SOAP-PP. \textbf{D.} Success rate for topN in predicting epitopes for ContactNet, AFM, and SOAP-PP.
}
\label{results}
\end{figure}

We find that ContactNet almost tripled the Top-1 success rate (see \ref{asessment}) compared to AFM and SOAP-PP (Table 1, Fig. \ref{results}A) for modeled antibodies. The Top-5 success rate was also significantly higher compared to both AFM and SOAP-PP. AFM had a better performance compared to SOAP-PP on our modeled antibodies dataset. Most of the hits selected by ContactNet in the Top-5 were of medium and acceptable accuracy (67\% and 33\%), while most of AFM models were of high accuracy (Fig. \ref{results}B). This can be explained by the low number of high-accuracy models sampled by the docking algorithms.

When tested  on 32 unbound antibody-antigen complexes from Benchmark 5.0 \cite{hwang2010protein}, ContactNet  achieved a higher performance with a success rate of 68\% and 75\% for Top-1 and Top-5 docking models, respectively (Fig. \ref{results}C). In contrast, SOAP-PP had a success rate of 20\% and 35\% and AFM had a success rate of 20\% and 40\%, for Top-1 and Top-5 respectively. The use of unbound antibodies should not affect AFM performance because only the sequence was used as an input. 

 We also tested ContactNet for epitope prediction using interface assessment criteria (see \ref{epitop}) \cite{lensink2010blind}. ContactNet reached a success rate of 50\% and 70\% for Top-1 and Top-10 docking models, respectively (Fig. \ref{results}D). In contrast, SOAP-PP had a success rate of 25\% (Top-1) and 50\% (Top-10) and AFM had a success rate of 38\% (Top-1) and 42\% (Top-5).

\section{Conclusions}
We present a novel deep learning architecture to address the problem of scoring antibody-antigen docking models. We train ContactNet for antibody-antigen complexes which present significant challenges, including the need to model antibodies and the lack of MSA that facilitates AFM prediction. ContactNet has significantly higher accuracy compared to state-of-the-art methods.

Several key factors contribute to the high performance of ContactNet. First, the compact 2D distogram representation for the protein and for the antibody-antigen interface makes the method invariant to transformations. Second, the network architecture is designed to simulate the biological process of protein-protein association where the interface contacts are formed by small complementary patches on the surfaces of the two proteins. The use of a problem-tailored architecture enables us to derive meaningful information out of the limited amount of data and create meaningful inductive bias. Finally, the residue level protein representation not only enables efficient training of the model using less GPU memory and runtime but also makes the network less sensitive to the differences between bound and unbound structures which often have different side-chain orientations.

\medskip

\bibliography{cites}

\begin{thebibliography}{31}
\providecommand{\natexlab}[1]{#1}
\providecommand{\url}[1]{\texttt{#1}}
\expandafter\ifx\csname urlstyle\endcsname\relax
  \providecommand{\doi}[1]{doi: #1}\else
  \providecommand{\doi}{doi: \begingroup \urlstyle{rm}\Url}\fi

\bibitem[Andrusier et~al.(2007)Andrusier, Nussinov, and
  Wolfson]{andrusier2007firedock}
Nelly Andrusier, Ruth Nussinov, and Haim~J Wolfson.
\newblock Firedock: fast interaction refinement in molecular docking.
\newblock \emph{Proteins: Structure, Function, and Bioinformatics}, 69\penalty0
  (1):\penalty0 139--159, 2007.

\bibitem[Baek et~al.(2021)Baek, DiMaio, Anishchenko, Dauparas, Ovchinnikov,
  Lee, Wang, Cong, Kinch, Schaeffer, et~al.]{baek2021accurate}
Minkyung Baek, Frank DiMaio, Ivan Anishchenko, Justas Dauparas, Sergey
  Ovchinnikov, Gyu~Rie Lee, Jue Wang, Qian Cong, Lisa~N Kinch, R~Dustin
  Schaeffer, et~al.
\newblock Accurate prediction of protein structures and interactions using a
  three-track neural network.
\newblock \emph{Science}, 373\penalty0 (6557):\penalty0 871--876, 2021.

\bibitem[Biewald(2020)]{wandb}
Lukas Biewald.
\newblock Experiment tracking with weights and biases, 2020.
\newblock URL \url{https://www.wandb.com/}.
\newblock Software available from wandb.com.

\bibitem[Cheng(2015)]{cheng2015single}
Yifan Cheng.
\newblock Single-particle cryo-em at crystallographic resolution.
\newblock \emph{Cell}, 161\penalty0 (3):\penalty0 450--457, 2015.

\bibitem[Comeau et~al.(2004)Comeau, Gatchell, Vajda, and
  Camacho]{comeau2004cluspro}
Stephen~R Comeau, David~W Gatchell, Sandor Vajda, and Carlos~J Camacho.
\newblock Cluspro: a fully automated algorithm for protein--protein docking.
\newblock \emph{Nucleic acids research}, 32\penalty0 (suppl\_2):\penalty0
  W96--W99, 2004.

\bibitem[Dai and Bailey-Kellogg(2021)]{dai2021protein}
Bowen Dai and Chris Bailey-Kellogg.
\newblock Protein interaction interface region prediction by geometric deep
  learning.
\newblock \emph{Bioinformatics}, 37\penalty0 (17):\penalty0 2580--2588, 2021.

\bibitem[Dong et~al.(2013)Dong, Fan, Schneidman-Duhovny, Webb, and
  Sali]{dong2013optimized}
Guang~Qiang Dong, Hao Fan, Dina Schneidman-Duhovny, Ben Webb, and Andrej Sali.
\newblock Optimized atomic statistical potentials: assessment of protein
  interfaces and loops.
\newblock \emph{Bioinformatics}, 29\penalty0 (24):\penalty0 3158--3166, 2013.

\bibitem[Eismann et~al.(2021)Eismann, Townshend, Thomas, Jagota, Jing, and
  Dror]{eismann2021hierarchical}
Stephan Eismann, Raphael~JL Townshend, Nathaniel Thomas, Milind Jagota, Bowen
  Jing, and Ron~O Dror.
\newblock Hierarchical, rotation-equivariant neural networks to select
  structural models of protein complexes.
\newblock \emph{Proteins: Structure, Function, and Bioinformatics}, 89\penalty0
  (5):\penalty0 493--501, 2021.

\bibitem[Evans et~al.(2021)Evans, O'Neill, Pritzel, Antropova, Senior, Green,
  {\v{Z}}{\'\i}dek, Bates, Blackwell, Yim, et~al.]{evans2021protein}
Richard Evans, Michael O'Neill, Alexander Pritzel, Natasha Antropova, Andrew~W
  Senior, Timothy Green, Augustin {\v{Z}}{\'\i}dek, Russell Bates, Sam
  Blackwell, Jason Yim, et~al.
\newblock Protein complex prediction with alphafold-multimer.
\newblock \emph{BioRxiv}, 2021.

\bibitem[Ferdous and Martin(2018)]{ferdous2018abdb}
Saba Ferdous and Andrew~CR Martin.
\newblock Abdb: antibody structure database—a database of pdb-derived
  antibody structures.
\newblock \emph{Database}, 2018, 2018.

\bibitem[Gray et~al.(2003)Gray, Moughon, Wang, Schueler-Furman, Kuhlman, Rohl,
  and Baker]{gray2003protein}
Jeffrey~J Gray, Stewart Moughon, Chu Wang, Ora Schueler-Furman, Brian Kuhlman,
  Carol~A Rohl, and David Baker.
\newblock Protein--protein docking with simultaneous optimization of rigid-body
  displacement and side-chain conformations.
\newblock \emph{Journal of molecular biology}, 331\penalty0 (1):\penalty0
  281--299, 2003.

\bibitem[Guest et~al.(2021)Guest, Vreven, Zhou, Moal, Jeliazkov, Gray, Weng,
  and Pierce]{guest2021expanded}
Johnathan~D Guest, Thom Vreven, Jing Zhou, Iain Moal, Jeliazko~R Jeliazkov,
  Jeffrey~J Gray, Zhiping Weng, and Brian~G Pierce.
\newblock An expanded benchmark for antibody-antigen docking and affinity
  prediction reveals insights into antibody recognition determinants.
\newblock \emph{Structure}, 29\penalty0 (6):\penalty0 606--621, 2021.

\bibitem[Hwang et~al.(2010)Hwang, Vreven, Janin, and Weng]{hwang2010protein}
Howook Hwang, Thom Vreven, Jo{\"e}l Janin, and Zhiping Weng.
\newblock Protein--protein docking benchmark version 4.0.
\newblock \emph{Proteins: Structure, Function, and Bioinformatics}, 78\penalty0
  (15):\penalty0 3111--3114, 2010.

\bibitem[Jumper et~al.(2021)Jumper, Evans, Pritzel, Green, Figurnov,
  Ronneberger, Tunyasuvunakool, Bates, {\v{Z}}{\'\i}dek, Potapenko,
  et~al.]{jumper2021highly}
John Jumper, Richard Evans, Alexander Pritzel, Tim Green, Michael Figurnov,
  Olaf Ronneberger, Kathryn Tunyasuvunakool, Russ Bates, Augustin
  {\v{Z}}{\'\i}dek, Anna Potapenko, et~al.
\newblock Highly accurate protein structure prediction with alphafold.
\newblock \emph{Nature}, 596\penalty0 (7873):\penalty0 583--589, 2021.

\bibitem[Kozakov et~al.(2006)Kozakov, Brenke, Comeau, and
  Vajda]{kozakov2006piper}
Dima Kozakov, Ryan Brenke, Stephen~R Comeau, and Sandor Vajda.
\newblock Piper: an fft-based protein docking program with pairwise potentials.
\newblock \emph{Proteins: Structure, Function, and Bioinformatics}, 65\penalty0
  (2):\penalty0 392--406, 2006.

\bibitem[Lensink and Wodak(2010)]{lensink2010blind}
Marc~F Lensink and Shoshana~J Wodak.
\newblock Blind predictions of protein interfaces by docking calculations in
  capri.
\newblock \emph{Proteins: Structure, Function, and Bioinformatics}, 78\penalty0
  (15):\penalty0 3085--3095, 2010.

\bibitem[Lensink et~al.(2007)Lensink, M{\'e}ndez, and
  Wodak]{lensink2007docking}
Marc~F Lensink, Ra{\'u}l M{\'e}ndez, and Shoshana~J Wodak.
\newblock Docking and scoring protein complexes: Capri 3rd edition.
\newblock \emph{Proteins: Structure, Function, and Bioinformatics}, 69\penalty0
  (4):\penalty0 704--718, 2007.

\bibitem[Lensink et~al.(2018)Lensink, Velankar, Baek, Heo, Seok, and
  Wodak]{lensink2018challenge}
Marc~F Lensink, Sameer Velankar, Minkyung Baek, Lim Heo, Chaok Seok, and
  Shoshana~J Wodak.
\newblock The challenge of modeling protein assemblies: the casp12-capri
  experiment.
\newblock \emph{Proteins: Structure, Function, and Bioinformatics},
  86:\penalty0 257--273, 2018.

\bibitem[London et~al.(2010)London, Raveh, Movshovitz-Attias, and
  Schueler-Furman]{london2010can}
Nir London, Barak Raveh, Dana Movshovitz-Attias, and Ora Schueler-Furman.
\newblock Can self-inhibitory peptides be derived from the interfaces of
  globular protein--protein interactions?
\newblock \emph{Proteins: Structure, Function, and Bioinformatics}, 78\penalty0
  (15):\penalty0 3140--3149, 2010.

\bibitem[M{\'e}ndez et~al.(2005)M{\'e}ndez, Leplae, Lensink, and
  Wodak]{mendez2005assessment}
Ra{\'u}l M{\'e}ndez, Rapha{\"e}l Leplae, Marc~F Lensink, and Shoshana~J Wodak.
\newblock Assessment of capri predictions in rounds 3--5 shows progress in
  docking procedures.
\newblock \emph{Proteins: Structure, Function, and Bioinformatics}, 60\penalty0
  (2):\penalty0 150--169, 2005.

\bibitem[Pal et~al.(2007)Pal, Chakrabarti, Bahadur, Rodier, and
  Janin]{pal2007peptide}
Arumay Pal, Pinak Chakrabarti, Ranjit Bahadur, Francis Rodier, and Jo{\"e}l
  Janin.
\newblock Peptide segments in protein-protein interfaces.
\newblock \emph{Journal of Biosciences}, 32\penalty0 (1):\penalty0 101--111,
  2007.

\bibitem[Pierce and Weng(2007)]{zrank}
Brian Pierce and Zhiping Weng.
\newblock Zrank: Reranking protein docking predictions with an optimized energy
  function.
\newblock \emph{Proteins: Structure, Function, and Bioinformatics}, 67\penalty0
  (4):\penalty0 1078--1086, 2007.
\newblock \doi{https://doi.org/10.1002/prot.21373}.
\newblock URL \url{https://onlinelibrary.wiley.com/doi/abs/10.1002/prot.21373}.

\bibitem[Renaud et~al.(2021)Renaud, Geng, Georgievska, Ambrosetti, Ridder,
  Marzella, R{\'e}au, Bonvin, and Xue]{renaud2021deeprank}
Nicolas Renaud, Cunliang Geng, Sonja Georgievska, Francesco Ambrosetti, Lars
  Ridder, Dario~F Marzella, Manon~F R{\'e}au, Alexandre~MJJ Bonvin, and Li~C
  Xue.
\newblock Deeprank: a deep learning framework for data mining 3d
  protein-protein interfaces.
\newblock \emph{Nature communications}, 12\penalty0 (1):\penalty0 1--8, 2021.

\bibitem[Ritchie and Kemp(2000)]{ritchie2000protein}
David~W Ritchie and Graham~JL Kemp.
\newblock Protein docking using spherical polar fourier correlations.
\newblock \emph{Proteins: Structure, Function, and Bioinformatics}, 39\penalty0
  (2):\penalty0 178--194, 2000.

\bibitem[Schneidman-Duhovny et~al.(2005)Schneidman-Duhovny, Inbar, Nussinov,
  and Wolfson]{schneidman2005patchdock}
Dina Schneidman-Duhovny, Yuval Inbar, Ruth Nussinov, and Haim~J Wolfson.
\newblock Patchdock and symmdock: servers for rigid and symmetric docking.
\newblock \emph{Nucleic acids research}, 33\penalty0 (suppl\_2):\penalty0
  W363--W367, 2005.

\bibitem[Singh et~al.(2020)Singh, Dauzhenka, Kundrotas, Sternberg, and
  Vakser]{singh2020application}
Amar Singh, Taras Dauzhenka, Petras~J Kundrotas, Michael~JE Sternberg, and
  Ilya~A Vakser.
\newblock Application of docking methodologies to modeled proteins.
\newblock \emph{Proteins: Structure, Function, and Bioinformatics}, 88\penalty0
  (9):\penalty0 1180--1188, 2020.

\bibitem[Veli{\v{c}}kovi{\'c} et~al.(2017)Veli{\v{c}}kovi{\'c}, Cucurull,
  Casanova, Romero, Lio, and Bengio]{velivckovic2017graph}
Petar Veli{\v{c}}kovi{\'c}, Guillem Cucurull, Arantxa Casanova, Adriana Romero,
  Pietro Lio, and Yoshua Bengio.
\newblock Graph attention networks.
\newblock \emph{arXiv preprint arXiv:1710.10903}, 2017.

\bibitem[Wang et~al.(2017)Wang, Sun, Li, Zhang, and Xu]{wang2017accurate}
Sheng Wang, Siqi Sun, Zhen Li, Renyu Zhang, and Jinbo Xu.
\newblock Accurate de novo prediction of protein contact map by ultra-deep
  learning model.
\newblock \emph{PLoS computational biology}, 13\penalty0 (1):\penalty0
  e1005324, 2017.

\bibitem[Wang et~al.(2020)Wang, Terashi, Christoffer, Zhu, and
  Kihara]{wang2020protein}
Xiao Wang, Genki Terashi, Charles~W Christoffer, Mengmeng Zhu, and Daisuke
  Kihara.
\newblock Protein docking model evaluation by 3d deep convolutional neural
  networks.
\newblock \emph{Bioinformatics}, 36\penalty0 (7):\penalty0 2113--2118, 2020.

\bibitem[Wang et~al.(2021)Wang, Flannery, and Kihara]{wang2021protein}
Xiao Wang, Sean~T Flannery, and Daisuke Kihara.
\newblock Protein docking model evaluation by graph neural networks.
\newblock \emph{Frontiers in Molecular Biosciences}, page 402, 2021.

\bibitem[Yin et~al.(2022)Yin, Feng, Varshney, and Pierce]{yin2022benchmarking}
Rui Yin, Brandon~Y Feng, Amitabh Varshney, and Brian~G Pierce.
\newblock Benchmarking alphafold for protein complex modeling reveals accuracy
  determinants.
\newblock \emph{Protein Science}, 31\penalty0 (8):\penalty0 e4379, 2022.

\end{thebibliography}

\medskip
\appendix
\section{Appendix}


\subsection{Dataset preparation}\label{Dataset}
 A dataset of $\sim$1,800 antibody-antigen complexes from the AbDb was used \cite{ferdous2018abdb}. Complexes with antigens larger than 700 or smaller than 20 amino acids were discarded, resulting in 1,214 complexes. Because this is a dataset of bound complex structures, we produced noisy unbound-like models for antibodies by modeling the antibodies via AFM, thus exposing the docking and the training to modeled structures. The complexes were divided into train and test using 97\% sequence identity cut-off for antibody sequences, The final train set included 790 complexes, with 85 additional complexes that were used for validation. The complexes were divided into train and test using 97\% sequence identity cut-off for antibody sequences.
 
 \label{DB5}We have also validated our method using 32 unbound antibody-antigen complexes from the docking Benchmark 5.0 {\cite{hwang2010protein}} by training a model that excludes all these complexes. This set contains unbound structures that enable us to test our method using more accurate antibody structures.

\subsection{Assessment criteria}
\label{asessment}
Each complex model is assessed for accuracy based on root mean square deviation (RMSD) from the correct structure, as used at CAPRI \cite{mendez2005assessment,lensink2007docking}. A docking model is considered acceptable if the ligand $C\alpha$ RMSD after superposition of the receptors is < 10$\textup{~\AA}$ or the interface $C\alpha$ RMSD is < 4$\textup{~\AA}$. A docking model is of medium accuracy if ligand $C\alpha$ RMSD is < 5$\textup{~\AA}$ or interface $C\alpha$ RMSD is < 2$\textup{~\AA}$. The success rate is the percentage of benchmark cases with at least one medium or acceptable accuracy model in the TopN predictions.

\subsection{Epitope prediction assessment}\label{epitop}
 Antigen interface assessment is performed using Recall and Precision criteria from CAPRI {\cite{lensink2010blind}}. Recall is the fraction of correctly predicted complex interface residues in the model and Precision is the fraction of residues in the model interface that are the actual complex interface residues. Interface residue is a residue whose solvent accessible area is lower in the complex (or model) than in the individual components. A predicted interface is considered correct if Recall and Precision are at least 0.5. Here, we calculated the Recall and Precision for the antigen only to estimate the accuracy of epitope identification.

\subsection{Training} ContactNet was trained end-to-end to predict binary classification based on the assumption that the distribution of correct and incorrect complexes is separable while optimizing the cross-entropy loss. We trained the model with the WADAM optimizer with a learning rate of 1e-4 that decays 100 times while training using the cosine decay as a learning rate scheduler with a weight of 5e-3 for the decay factor. The batch size contained 52 docking models, randomly selected from different complexes for better generalization. One epoch was defined as 2,000 batches with ~100,000 docking models per epoch. The network was trained for 160 epochs. The major challenge in the training process was the unbalanced nature of the data. Specifically, there were hundreds of thousands of negative complexes compared to a small number of positive ones. To cope with this challenge each batch is composed of 25\% of the positive, acceptable accuracy docking models and 75\% of the negative, incorrect ones. We used the W\&B platform for experiment tracking {\cite{wandb} }. and trained the model on a single GPU 2080RTX for 20 hours.

\subsection{Sampling antibody-antigen complexes} \label{sampling} The antibodies were docked to antigens using the antibody-antigen docking protocol of PatchDock. PatchDock is an efficient geometric rigid docking method that maximizes shape complementarity {\cite{schneidman2005patchdock}}.Here we used a higher sampling precision to generate a higher fraction of “positives”, eg. acceptable accuracy models, for training, typically generating ~100,000 docking models. These models were scored and re-ranked using SOAP-PP statistical potential {\cite{dong2013optimized}}. For training, we used 2,500 top-scoring models along with up to 50 positives irrespective of their ranking. For validation, we have used 2,500 top-scoring docking models based on SOAP-PP ranking.

\subsection{Scoring function}
To visualize the discrimination properties of the two scoring functions and the AFM, we examined the test set funnel plots . We found that ContactNet is highly selective, classifying most of the docking models as strictly negative (score = 0.0).When positives are correctly identified (Fig. S1, PDB 1KIP). In contrast, the funnels of SOAP-PP have a wider spread of docking models across the range of score values (y-axis). In some cases, ContactNet fails to identify accurate complexes (Fig. S1, PDB 1IAI). However, even when it fails, the epitope is often correctly predicted, but the antibody orientation is incorrect (Fig. S1, PDB 1IAI).

\begin{figure}
\centering
\includegraphics[scale=0.5]{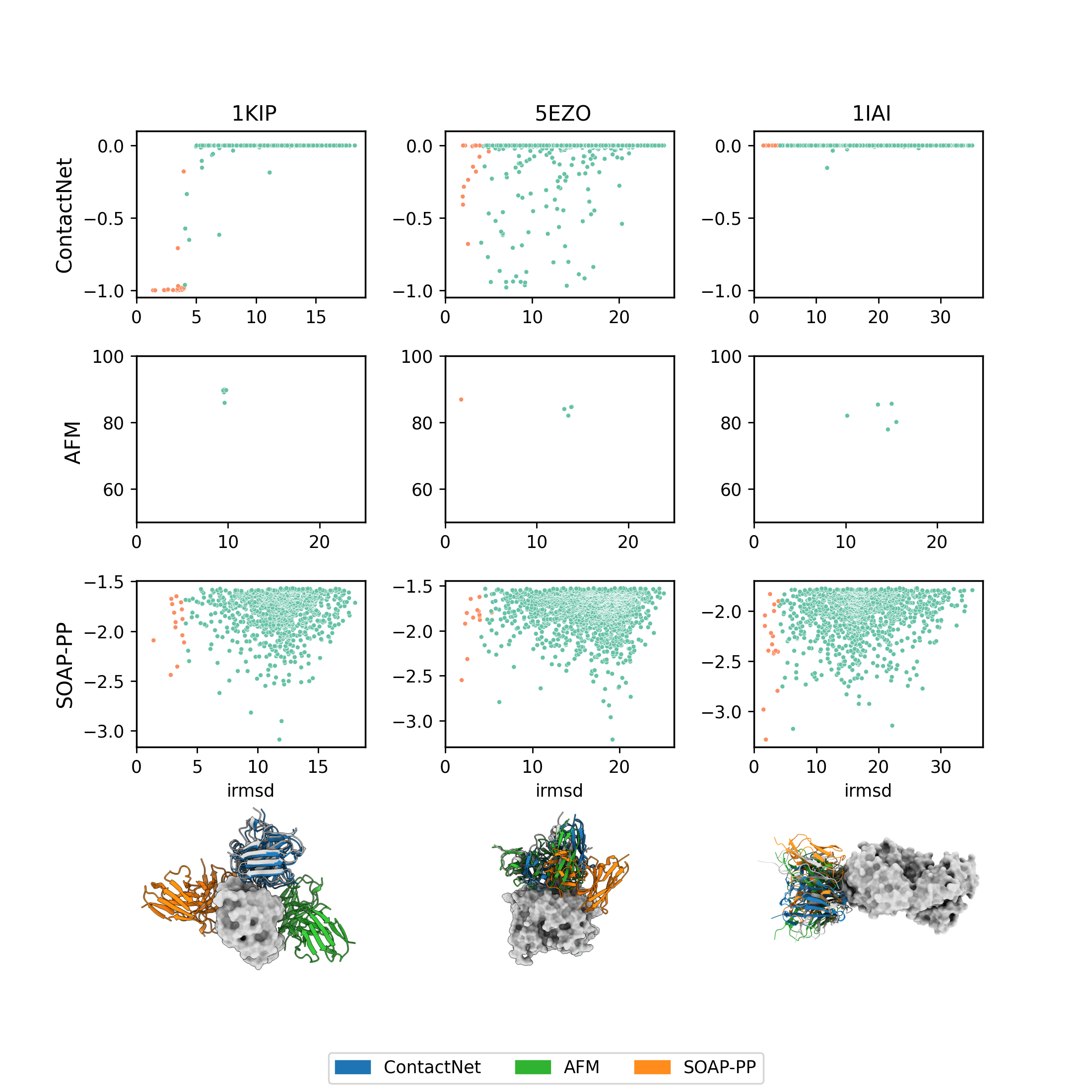}
\caption{\textbf{Figure 1S. Funnels for scoring functions.} \textbf{A.} ContactNet, AFM, and SOAP-PP funnels as a function of interface RMSD for the three cases from the test set (1KIP, 5EZO, and 5IAI). Each dot corresponds to a docking model. Acceptable accuracy models are shown in red. \textbf{B}. The top scoring models for each scoring function (ContactNet - blue, AFM - green, SOAP-PP- orange) are shown at the bottom (overlaid on the gray X-ray structure).}
\label{results}
\end{figure}

\end{document}